%% file: main.tex
\pdfoutput=1

\documentclass[10pt,conference,letterpaper]{IEEEtran}
\usepackage{url}
\usepackage{amsmath}
\usepackage{caption}
\usepackage{xcolor}
\usepackage{enumerate}
\usepackage{mathtools}
\usepackage{amsthm}
\usepackage{multirow}
\usepackage{longtable}
\ifCLASSOPTIONcompsoc
  \usepackage[caption=false,font=normalsize,labelfont=sf,textfont=sf]{subfig}
\else
  \usepackage[caption=false,font=footnotesize]{subfig}
\fi
\usepackage{graphicx}

\newlength{\oldtabcolsep}

\newcommand{\Name}{Fathom}
\newcommand{\wkl}[1]{{\tt #1}}

\begin{document}

\title{{\Name}: Reference Workloads for\\ Modern Deep Learning Methods}

\author{%
\IEEEauthorblockN{Robert Adolf, Saketh Rama, Brandon Reagen, Gu-Yeon Wei, and David Brooks}%
\IEEEauthorblockA{Harvard University}%
}
\maketitle

\input{abstract}
\input{motivation_table}
\section{Introduction}\label{sec:intro}\input{intro}

\section{Motivation: The Tip of the Iceberg}\label{sec:motivation}\input{motivation}

\section{Design and Implementation Criteria}\label{sec:methodology}\input{methodology}
\input{workloads_table}
\section{The {\Name} Workloads}\label{sec:workloads}\input{workloads}

\section{Understanding the Performance Characteristics of Deep Learning Workloads}\label{sec:analysis}\input{analysis}
\section{Related work}\label{sec:related_work}\input{related_work}
\section{Conclusion}\label{sec:conclusion}\input{conclusion}
\section*{Acknowledgements}
This work was partially supported by C-FAR, one of six centers of STARnet, a Semiconductor Research Corporation program sponsored by MARCO and DARPA. The work was also supported in part by DARPA under Contract \#: HR0011-13-C-0022. This research was, in part, funded by the U.S. Government. The views and conclusions contained in this document are those of the authors and should not be interpreted as representing the official policies, either expressed or implied, of the U.S. Government.

\IEEEtriggeratref{31}
\bibliographystyle{abbrv}
\bibliography{dnn.bib}

\end{document}

%% file: abstract.tex

\begin{abstract}
Deep learning has been popularized by its recent successes on challenging artificial intelligence problems.
One of the reasons for its dominance is also an ongoing challenge: the need for immense amounts of computational power.
Hardware architects have responded by proposing a wide array of promising ideas, but to date, the majority of the work has focused on specific algorithms in somewhat narrow application domains.
While their specificity does not diminish these approaches, there is a clear need for more flexible solutions.
We believe the first step is to examine the characteristics of cutting edge models from across the deep learning community.

Consequently, we have assembled Fathom: a collection of eight archetypal deep learning workloads for study.
Each of these models comes from a seminal work in the deep learning community, ranging from the familiar deep convolutional neural network of Krizhevsky et al., to the more exotic memory networks from Facebook's AI research group.
Fathom has been released online, and this paper focuses on understanding the fundamental performance characteristics of each model.
We use a set of application-level modeling tools built around the TensorFlow deep learning framework in order to analyze the behavior of the Fathom workloads.
We present a breakdown of where time is spent, the similarities between the performance profiles of our models, an analysis of behavior in inference and training, and the effects of parallelism on scaling.
\end{abstract}

%% file: motivation_table.tex

\newcommand{\x}{$\times$}
\setlength\oldtabcolsep{\tabcolsep}
\setlength\tabcolsep{3pt}
\begin{table*}
    \centering
    \caption{Recent Architecture Research in Deep Learning}{There are a wide variety of deep learning problems which are largely untouched by current work.}
    \vspace{1ex}
    \label{table:motivation}
    \begin{tabular}{c|c|c|c|c|c|c|c|c|c|c|c|c|c|c|c|c|c|c}\hline
        \hline
        Category & Feature
        &\cite{earlyCNN}
        &\cite{chen2012-benchnn}
        &\cite{diannao}
        &\cite{chen-2014}
        &\cite{eyeriss}
        &\cite{reram}
        &\cite{shidiannao}
        &\cite{eie}
        &\cite{gin}
        &\cite{pud}
        &\cite{ovtcharov2015accelerating}
        &\cite{minerva}
        &\cite{isaac}
        &\cite{cortex}
        &\cite{ffts}
        &\cite{jasonCNNs}
        &{\Name} \\
        \hline
        \hline
        \multirow{3}{*}{Neuronal Style}
        & Fully-connected        &\x&\x&\x&  &  &\x&\x&\x&\x&\x&\x&\x&\x&  &\x&  &\x \\
        & Convolutional          &\x&  &\x&\x&\x&\x&\x&  &\x&  &\x&  &\x&  &  &\x&\x \\
        & Recurrent              &  &  &  &  &  &  &  &\x&  &  &  &  &  &\x&  &  &\x \\
        \hline
        Layer Depth & (Maximum)  & 4& 4& 3& 3& 5&16& 7& 3&13& 6& 9& 4&26& 2& 5& 5&34 \\
        \hline
        \multirow{4}{*}{Learning Task}
        & Inference              &\x&\x&\x&\x&\x&\x&\x&\x&\x&\x&\x&\x&\x&\x&\x&\x&\x \\
        & Supervised             &  &\x&  &\x&  &  &  &  &\x&\x&\x&\x&  &\x&  &  &\x \\
        & Unsupervised           &  &  &  &  &  &  &  &  &  &  &  &  &  &  &  &  &\x \\
        & Reinforcement          &  &  &  &  &  &  &  &  &  &  &  &  &  &  &  &  &\x \\
        \hline
        \multirow{4}{*}{Application Domain}
        & Vision                 &\x&  &\x&\x&\x&\x&\x&\x&\x&\x&\x&\x&\x&  &  &\x&\x \\
        & Speech                 &  &  &  &  &  &  &  &\x&\x&  &  &  &  &  &  &  &\x \\
        & Language Modeling      &  &  &  &  &  &  &  &\x&\x&  &  &\x&  &\x&  &  &\x \\
        & Function Approximation &  &\x&  &  &  &  &  &  &  &  &  &  &  &  &\x&  &\x \\
        \hline
    \end{tabular}
    \vspace{-2ex}
\end{table*}
\setlength\tabcolsep{\oldtabcolsep}

%% file: intro.tex

The current surge in deep learning research can largely be attributed to three complementary trends: steady progress in clever algorithms, the availability of massive amounts of data, and the ever-increasing computational power of modern hardware.
These trends form a virtuous cycle: improvement in one area drives demand for improvement in the others, drawing resources and bright minds, eventually yielding new developments and sustaining the cycle.
From a hardware architect's point of view, there are plenty of opportunities.
There is an insatiable computational demand from existing deep learning methods, and the ability of machine learning experts to discover new methods is directly tied to the training speed of their models.
Additionally, a variety of new platforms are emerging as deep learning application targets, including mobile platforms, autonomous systems, and smart devices.
Each presents a unique set of constraints, and all require novel hardware to move forward.

The architecture community has begun to respond in force:
A host of new hardware designs have been proposed \cite{diannao,eyeriss,eie,minerva}, and several are beginning to transition into production \cite{jouppi2016tpu,ovtcharov2015accelerating}.
We believe this trend is just the beginning.
As deep learning techniques continue to be extended into new application areas and more challenging problems, the scope for novel computational support will only grow.
To date, however, the majority of the research has focused on exploiting a narrow set of algorithms.
While the results have been impressive, this sharp focus has left many application areas underserved.
A broader approach will require a more comprehensive understanding of the workload characteristics across the entire application space.

To address this need and enable research into more versatile architectures, we have assembled {\Name}, a set of workloads drawn from across the community to serve as exemplars for the many approaches practitioners take to solving dissimilar problems.
These models are meant to represent the state of the art in deep learning.
They are not toy problems.
Most are taken directly from top-tier research venues and have either set new accuracy records on competitive datasets (e.g., image classification) or pushed the limits of our understanding of deep learning (e.g., agent planning).

To facilitate adoption in the architecture community, we also provide an overview of the performance behavior for the {\Name} workloads.
We use a custom, high-level analysis framework built around TensorFlow to highlight execution characteristics in the context of the models themselves.
Specifically, we focus on four areas: identifying the types of operations which dominate execution time, measuring similarity between deep learning models, comparing the performance differences between training and inference, and understanding the effects of parallel scalability.

This paper makes the following contributions:
\begin{itemize}
\item{We demonstrate the need for a broader variety of deep learning workloads by looking at the models used in recent architecture literature. (Section~\ref{sec:motivation})}
\item{We introduce {\Name}, a set of reference implementations of state-of-the-art deep learning models. (Sections~\ref{sec:methodology},~\ref{sec:workloads})}
\item{We provide a quantitative analysis of the fundamental computational characteristics of the {\Name} workloads. (Section~\ref{sec:analysis})}
\end{itemize}

%% file: motivation.tex

Deep learning algorithms are inveterate computational hogs.
As soon as a more powerful hardware platform is released, learning models invariably expand to fill the available resources.
This is not due to inefficiency: these models are attacking human-level artificial intelligence problems that have never been solved, and computation is a major factor in their ability to make inroads on these challenges.
Baidu researchers have estimated that a single training run of a modern deep learning algorithm requires at least 10 exaflops to converge~\cite{amodei2015-deepspeech2}.
Computational power also drives innovation.
Jeff Dean recently noted that research productivity is inversely proportional to the turnaround time of a deep learning experiment: while 1--4 days is ``tolerable'', ``progress stalls'' on the order of weeks, and month-long experiments are not worth running~\cite{dean2015_baylearn}.

As a direct result, there has been a surge of research in hardware support for deep learning.
The prospective gains these systems offer are tantalizing:
improvements in performance or power on the order of 10--100$\times$ have been suggested by a number of groups.
That level of improvement could turn previously intractable research into overnight jobs, open up new classes of learning algorithms, and enable deep learning solutions be deployed in even the most computationally constrained environments.

Unfortunately, there is a lingering mismatch which is standing in the way:
much of the research in the architecture community is focused on accelerating a relatively narrow set of deep learning techniques.
There is undeniable benefit for these chosen problem domains, but there is a wider space of applications for which solutions have yet to be found.
We can see this phenomenon by examining the deep learning techniques found in the published literature.
Table~\ref{table:motivation} presents a survey of 16 recent papers from top-tier architecture conferences.
A cursory look shows that while a growing body of work already exists, much of it falls into similar veins.
Nearly half of these papers evaluate the same neural network (the well-known CNN from Krizhevsky et al.~\cite{krizhevsky2012-alexnet}).
While this has the advantage of commensurability, it is not by itself representative of the larger deep learning community.
Similar trends apply to learning modalities:
while the architecture community seems to be comfortable with supervised learning tasks (and the inference-only version of them), we were unable to find any recent hardware work in support of unsupervised or reinforcement deep learning problems.
Finally, despite being the state of the art for both speech recognition and language modeling, recurrent neural networks appeared just twice: a heavily modified version of Karpathy and Li's NeuralTalk work~\cite{karpathy2015-neuraltalk} in Han et al.~\cite{eie} and a slightly-dated restricted Boltzmann machine in Thomas et al.~\cite{cortex}.

This is an opportunity.
It signifies that, far from being ``mined out,'' we are just scratching the surface of what can be done in this area.
In particular, we see two promising avenues for exploration:
first, there is a clear opening for new hardware designs to support underserved deep learning methods.
Unsupervised and reinforcement learning have largely been untouched by the architecture community, as are recurrent networks.
Second, there is a need for flexible architectures which can offer better performance or efficiency on a broad set of deep learning tasks.
We see this paper as providing one tool for starting down those paths: a set of standard, verified, reference workloads which represent the current state of the art in deep learning methods.

%% file: methodology.tex

\newcommand{\criterion}[1]{\subsection{#1}}

\criterion{Choose meaningful models}
The first question is how to select the right models to reflect the state of the art.
We believe this set of workloads should have three properties: representativeness, diversity, and impact.
The first is clear: our choices should reflect the best of what the deep learning community has come up with.
Since there are easily dozens of models which could rightly claim this status, the need to keep the size of the set to a manageable number implies a need for diversity: each model should bring something unique to the table.
Finally, impact is the degree to which a particular technique has changed the landscape of deep learning research.
Since we cannot predict the state of the field in five years, we instead try to choose methods which have imparted fundamental lessons to the work that came after---lessons that will continue to be relevant to future models.

One important note is that we do \emph{not} attempt to capture pre- or post-processing steps on either the input data or the model's output.
Many models depend on preprocessing for robustness and generality, but these techniques fall outside of the scope of this paper and are handled adequately by existing benchmark suites.

\criterion{Faithfully reproduce the original work}
This leads to the question of implementation.
We are lucky in that so much of the research in the deep learning community has taken place in the open.
The vast majority of research papers make a concerted effort to describe their topology and hyperparameters in the name of reproducibility, and many groups have opted to place reference implementations online.
This is because reproducing deep learning experiments can be challenging; small changes in the choice of tuning parameters or data preparation can lead to large differences in outcome.
So in crafting implementations for our chosen models, we adopt an existing implementation if one is available, translate one from a different source language if not, or create and validate an implementation using descriptions from the original paper.
All eight Fathom workloads were rewritten to adhere to a standard model interface, expose information to our profiling tools, and remove preprocessing or extraneous logging operations.

Similar reproducibility concerns apply to data sources as well.
Whenever possible, we run our workloads using the same training and test data as the original paper.
In cases where this is not possible (i.e., where the data is proprietary), we choose another dataset used by similar papers in the same area.
So, for instance, since we cannot train Deep Speech on Baidu's massive, private collection of recorded utterances, we substitute the widely cited TIMIT corpus~\cite{garofolo1993-timit}.

\criterion{Leverage a modern deep learning framework}
One particularly dramatic shift in the deep learning community in the last decade has been the widespread adoption of high-level programming models.
These frameworks provide two main benefits:
first, they abstract the underlying hardware interface away from the programmer.
High-performance code is difficult and time-consuming to write, regardless of whether it is vectorized C++ on a CPU or CUDA on a GPU.
Second, they provide libraries of useful kernels contributed by a variety of people, which acts as a productivity multiplier, especially in a fast-moving environment.
These frameworks have changed the development landscape, largely for the better, and it is not possible to create a realistic set of deep learning workloads without taking them into account.

Unfortunately, using a high-level framework also raises questions.
Primarily, to what extent does the choice of framework matter and which one should we choose?
The answer to the former question is, perhaps surprisingly, not all that much.
This is in large part due to a case of convergent evolution amongst the most popular libraries.
Consider four most widely used frameworks:
Torch~\cite{torch}, a general machine learning library written in Lua;
TensorFlow~\cite{tensorflow}, the dataflow-based second generation of Google's DistBelief system~\cite{distbelief};
Theano~\cite{theano}, a symbolic mathematics package originally from Universit\'e de Montr\'eal;
and Caffe~\cite{jia2014caffe}, Berkeley's deep learning library with a JSON frontend and C++ backend.
All four share very similar high-level structure, and the authors seem to have arrived at many of the same design decisions:
\begin{itemize}
\item{All use a simple front-end specification language, optimized for productivity.}
\item{All use highly-tuned native backend libraries for the actual computation. For NVidia GPUs, all four leverage the cuDNN package~\cite{cudnn}.}
\item{Most use an application-level, compiler-esque optimizer (Torch does not).}
\item{Most are declarative, domain-specific languages (only Torch is imperative).}
\item{Most provide support for automatic differentiation (Caffe's layers are hand-implemented).}
\item{All have some notion of a fundamental building block or primitive operation: TensorFlow and Theano call them operations, Torch calls them modules, and Caffe uses layers.}
\end{itemize}

There will undeniably be differences in the performance characteristics of a model implemented in two of these libraries.
However, essential traits are retained.
First, the task of writing a model in any of these frameworks consists of assembling a pipeline of primitive operations, and most of these primitives have direct analogues in other frameworks.
This similarity is so strong, in fact, that automatic translators exist for taking a model in one framework and emitting one in another~\cite{loadcaffe,caffe-tensorflow}
and wrapper interfaces exist for automatically generating a model in several output frameworks~\cite{keras}.
Second, the performance of models written in these languages are largely dominated by the primitive operations they contain, not the overhead of the framework which contains them.
This means that regardless of how the neural network is constructed, the performance characteristics will depend on the number, type, and organization of these operations (we show quantitative evidence of this in Section~\ref{subsec:measurement}).

Ultimately, while an argument could be made for choosing any one of these frameworks, the decision is somewhat arbitrary because of the similarities.
We implement both our reference workloads and the analysis tools used in this paper in TensorFlow, but the models would not be difficult to port to another framework, and our tools are designed to be agnostic.

%% file: workloads_table.tex

\setlength\oldtabcolsep{\tabcolsep}
\setlength\tabcolsep{3pt}
\begin{table*}
\caption{The {\Name} Workloads}
\begin{center}
\begin{tabular}{c|c|c|c|c|c|p{0.35\textwidth}}
Model Name & Year and Ref& Neuronal Style & Layers & Learning Task & Dataset & Purpose and Legacy \\
\hline
\hline
\wkl{seq2seq} &2014 \cite{sutskevar2014-seq2seq}& Recurrent & 7 & Supervised  & WMT-15 \cite{callisonburch2009-wmt} & Direct language-to-language sentence translation. State-of-the-art accuracy with a simple, language-agnostic architecture. \\
\hline
\wkl{memnet} &2015 \cite{sukhbaatar2015-end2end_memory}& Memory Network & 3 & Supervised & bAbI \cite{weston2015-babi} & Facebook's memory-oriented neural system. One of two novel architectures which explore a topology beyond feed-forward lattices of neurons. \\
\hline
\wkl{speech} &2014 \cite{hannun2014-deep_speech}& Recurrent, Full & 5 & Supervised & TIMIT \cite{garofolo1993-timit} & Baidu's speech recognition engine. Proved purely deep-learned networks can beat hand-tuned systems. \\
\hline
\wkl{autoenc} &2014 \cite{kingma2014-variational}& Full & 3 & Unsupervised  &  MNIST \cite{lecun1998-mnist} & Variational autoencoder. An efficient, generative model for feature learning.\\
\hline
\wkl{residual} &2015 \cite{he2015-deep_residual_learning}& Convolutional & 34 & Supervised & ImageNet \cite{deng2009-imagenet} & Image classifier from Microsoft Research Asia. Dramatically increased the practical depth of convolutional networks. ILSVRC 2015 winner.\\
\hline
\wkl{vgg} &2014 \cite{simonyan2014-vgg}& Convolutional, Full & 19 & Supervised & ImageNet \cite{deng2009-imagenet} & Image classifier demonstrating the power of small convolutional filters. ILSVRC 2014 winner. \\
\hline
\wkl{alexnet} &2012 \cite{krizhevsky2012-alexnet}& Convolutional, Full & 5 & Supervised & ImageNet \cite{deng2009-imagenet} & Image classifier. Watershed for deep learning by beating hand-tuned image systems at ILSVRC 2012. \\
\hline
\wkl{deepq} &2013 \cite{mnih2013-deepq}& Convolutional, Full & 5 & Reinforcement & Atari ALE \cite{bellemare2013-arcade} & Atari-playing neural network from DeepMind. Achieves superhuman performance on majority of Atari2600 games, without any preconceptions. \\
\hline
\end{tabular}
\end{center}
\label{default}
\end{table*}%
\setlength\tabcolsep{\oldtabcolsep}

%% file: workloads.tex

\subsection*{Sequence-to-Sequence Translation}
\wkl{seq2seq} is recurrent neural network for solving machine translation~\cite{sutskevar2014-seq2seq}.
The technique, developed at Google in 2014, uses a multi-layer pipeline of long short-term memory (LSTM) neurons to extract the meaning of a sentence and then re-emit it into another language.
The core neural network is comprised of three 7-neuron layers through which word tokens flow unidirectionally.
The model also leverages an attention-based model for keeping track of context in the original sentence~\cite{bahdanau2015-attention_encoder}.
Sequence-to-sequence translation succeeded in achieving best-of-breed accuracy, but its impact is largely derived from its elegance and flexibility.
It is a canonical example of a recurrent ``encoder-decoder'' model, a technique which transforms an input into a vector in high-dimensional space, called an embedding.

\subsection*{End-to-End Memory Networks}
Memory networks~\cite{weston2014memory} are one of two recent efforts to decouple state from structure in a neural network.
The development of memory networks stemmed from the difficulty that stateful neurons have in capturing long-range dependencies.
Facebook's AI research group solved this problem by joining an indirectly addressable memory with a neural network, resulting in a model which can explicitly store and recall information.
End-to-end memory networks~\cite{sukhbaatar2015-end2end_memory} are an extension which removes the need for type annotations on inputs and dramatically streamlines training.
The bAbI question-answer dataset is a natural language reasoning problem, where a model must make simple logical deductions from an unordered sequence of statements.

\subsection*{Deep Speech}
Deep Speech was Baidu Research's attempt at a scalable speech recognition model~\cite{hannun2014-deep_speech}.
The model is five fully-connected layers of 2048 neurons each with one bidirectional recurrent layer.
Deep Speech is a pure deep learning algorithm, in that it uses spectrograms directly as inputs and learns to transcribe phonemes (as opposed to using a hand-tuned acoustic model or HMM as a preprocessing stage).
Its connectionist temporal classification (CTC) loss function can learn from unsegmented data, significantly reducing the cost of producing training data~\cite{graves2006connectionist}.
Deep Speech was also notable for its emphasis on efficiency: the researchers explicitly designed the model to perform well on a GPU.
We implemented the Deep Speech architecture using smaller window and embedding sizes to account for differences in the TIMIT dataset~\cite{garofolo1993-timit}.

\subsection*{Variational Autoencoder}
Autoencoders are a flexible, unsupervised model often used for dimensionality reduction, feature extraction, or generating data~\cite{hinton2006reducing}.
The fundamental assumption is that there exists a compact representation of all realistic inputs (called an embedding) which can be used to both analyze and synthesize data.
Variational autoencoders, invented by Kingma and Welling in 2013, make a statistical assumptions about the properties of this embedding in order to learn to efficiently reconstruct their inputs~\cite{kingma2014-variational}.
These models are somewhat unique in that they require stochastic sampling as part of inference, not just training.

\subsection*{Residual Networks}
Residual networks were a landmark in enabling very deep neural networks~\cite{he2015-deep_residual_learning}.
Researchers at Microsoft Research Asia in 2015 confronted the phenomenon where increasing the depth of a model degrades both training \emph{and} validation error.
Their solution was to add additional identity connections across every pair of convolutional layers, effectively training these layers on the difference between its input and output.
This tactic enabled them to train models of over 150 layers deep, almost seven times larger than the previous state of the art, and it won them all five 2015 ILSVRC competitions tracks.

\subsection*{VGG-19}
\wkl{vgg} is an implementation of the 19-layer convolutional network developed by the visual geometry group at Oxford~\cite{simonyan2014-vgg}.
The success of AlexNet inspired deeper convolutional networks, and VGG was one such offspring.
The key insight by Simonyan and Zisserman was that more layers of smaller convolutional filters are easier to train.
This technique both improved accuracy (winning the ILSVRC localization task and placing 2nd in the classification task against a far more complex Google entry) and dramatically reduced the number of learnable parameters.

\subsection*{AlexNet}
AlexNet~\cite{krizhevsky2012-alexnet} was a watershed event for the deep learning community.
While now overshadowed by more advanced models, the original model made several significant contributions.
Foremost, it demonstrated that an automatically trained neural network could surpass hand-tuned image classifiers by a substantial margin.
It also introduced dropout as a regularization mechanism and showcased the computational power of GPUs.
While a large portion of the architecture community is already working with AlexNet, its inclusion adds both continuity (allowing some degree of commensurability with prior work) as well as a reference point for the other models in {\Name}.

\subsection*{Deep Reinforcement Learning}
DeepMind startled the AI community in 2013 with a deep reinforcement learning system that learned to win dozens of Atari games solely from pixels and scores, sometimes beating human experts \cite{mnih2013-deepq, mnih2015human}.
Unlike supervised algorithms, the deep Q learning algorithm improves its chosen actions as it receives in-game feedback, not by observing perfect play.
The heart of the method is a convolutional network which selects actions using 2-3 convolutional layers and 2-3 dense layers.
The model circumvented historical difficulties in extending neural networks to decoupled feedback with innovative strategies such as experience replay.
We leverage the same Atari emulation environment which powered the original implementation, the Arcade Learning Environment~\cite{bellemare2013-arcade}.

%% file: analysis.tex

These models are intended to be a tool for architects, and proper tools require skill and understanding to wield effectively.
This section is meant to build the foundation of that understanding in two ways:
First, we want to give architects an intuition about the behavior of deep learning workloads. It is important to understand, for instance, where time is actually spent, and what the relationships are between a given model and the hardware it runs on.
Second, we want to supply a quantitative baseline on which the community can build. There is a good deal of folklore surrounding deep learning, and numbers are a good way to begin to dispel some of that.

\subsection{Measurement and analysis in a deep learning framework}\label{subsec:measurement}
Working with a high-level framework like TensorFlow is a double-edged sword.
On one hand, it is a complex, dynamically-compiled, dataflow-oriented runtime system, so it causes problems for conventional analysis tools.
Profiling at the scripting-language level with a tool like cPython is difficult because TensorFlow is declarative, so all of the actual computation is deferred.
Low-level profiling (including call-path profiling) or hardware counter instrumentation with a tool like Linux's perf can provide detailed performance information, but it loses all connection to the original model: a call to a low-level library routine cannot easily be assigned to a particular layer, for instance, because those layers only exist as internal data structures.
On the other hand, TensorFlow itself makes an attractive platform for measurement.
The primitive operations which are used to construct a model are ideal targets for instrumentation, and a tool built around the framework already has easy access to model information like layers and parameters, so ascribing runtime behavior to model features is straightforward.
Additionally, TensorFlow (like many other popular deep learning frameworks) comes with some degree of built-in tracing support.
We leverage all of these features to build a set of characterization tools which focus on capturing performance information at the model level, and we use them for all of the analyses described in this paper.

Because we use operations as the primary abstraction for understanding the performance of the {\Name} models, it is worth spending time to explain them more thoroughly.
(Although we will only describe TensorFlow here, the principles are also applicable to the other three popular libraries mentioned in Section~\ref{sec:methodology} and to most production-quality deep learning frameworks in general.)
An operation is a node in the coarse-grained dataflow graph that defines a TensorFlow model.
It is implemented as a Python function which instructs the framework to build that node, as well as a C++ function which either performs the calculation or calls down to a low-level library to do so (either the Eigen linear algebra package on a CPU or a CUDA library like cuBLAS or cuDNN on a GPU).
Operations are the smallest schedulable unit in the TensorFlow runtime, and they double as the mechanism behind its symbolic auto-differentiation support.
Examples include functions like 2D matrix-matrix multiplication (MatMul), elementwise tensor exponentiation (Pow), or specialized functions like sampling from a normal distribution (StandardRandomNormal) or computing the loss function for a particular optimization algorithm (CrossEntropy).

\begin{figure}[h!]
\begin{center}
\includegraphics[width=\columnwidth]{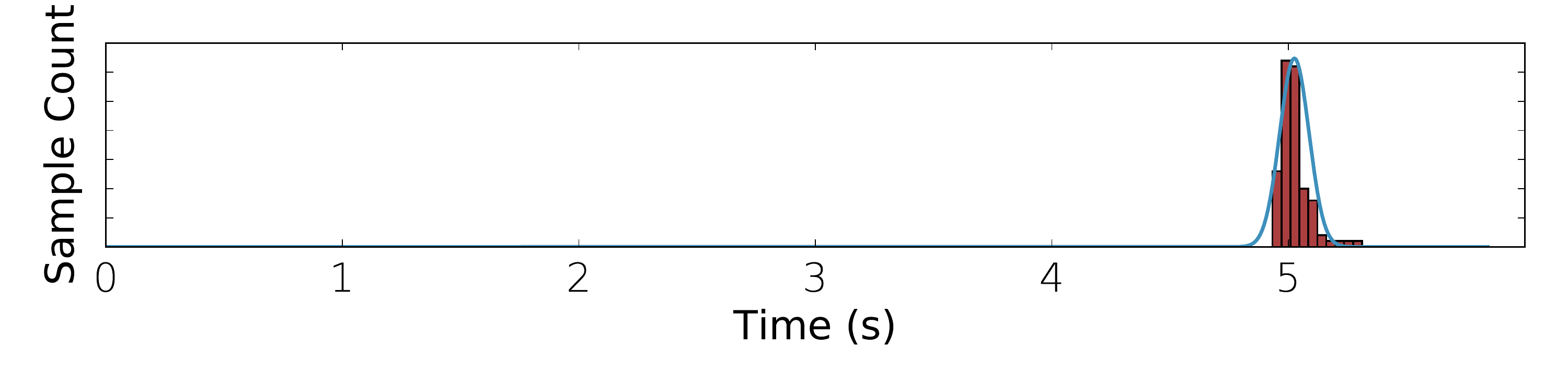}
\vspace{-3ex}
\caption{Sampling operations across the life of a program shows their execution time is stationary has low variance.}
\label{fig:stationarity}
\vspace{-1ex}
\end{center}
\end{figure}

Decomposing models into their component operations is convenient from a performance measurement standpoint.
First, operations tend to have stable, repeatable behavior across the life of a program.
Most deep learning models use some variant of gradient descent and backpropagation for optimization, so programs are naturally separable on update-step boundaries (also called minibatches for many training problems) or between inferences.
Sampling the execution time of operations across many steps allows us to quantify stability, and Figure~\ref{fig:stationarity} shows that this distribution is stationarity and has low variance.
Second, most deep learning models are dominated by the time spent inside their operations.
Our measurements reveal that inter-operation overhead is minimal in TensorFlow: typically less than 1--2\% of the total runtime is spent outside of operations in our workloads.

Finally, our experiments are carried out on a 4GHz Skylake i7-6700k with 32GB RAM or an NVidia GeForce GTX 960, running TensorFlow v0.8, CUDA 7.0, and cuDNN 6.5-v2.
While GPUs are more popular for the performance they provide, many frameworks (TensorFlow included) have incomplete support for all operations, and the fall-back behavior is to run unsupported operations on the CPU, splitting execution across the PCI bus.
This causes crippling performance problems, so in order to avoid analysis artifacts, we opt for running most experiments on a CPU.

\begin{figure}[htbp]
\begin{center}
\includegraphics[width=.95\columnwidth]{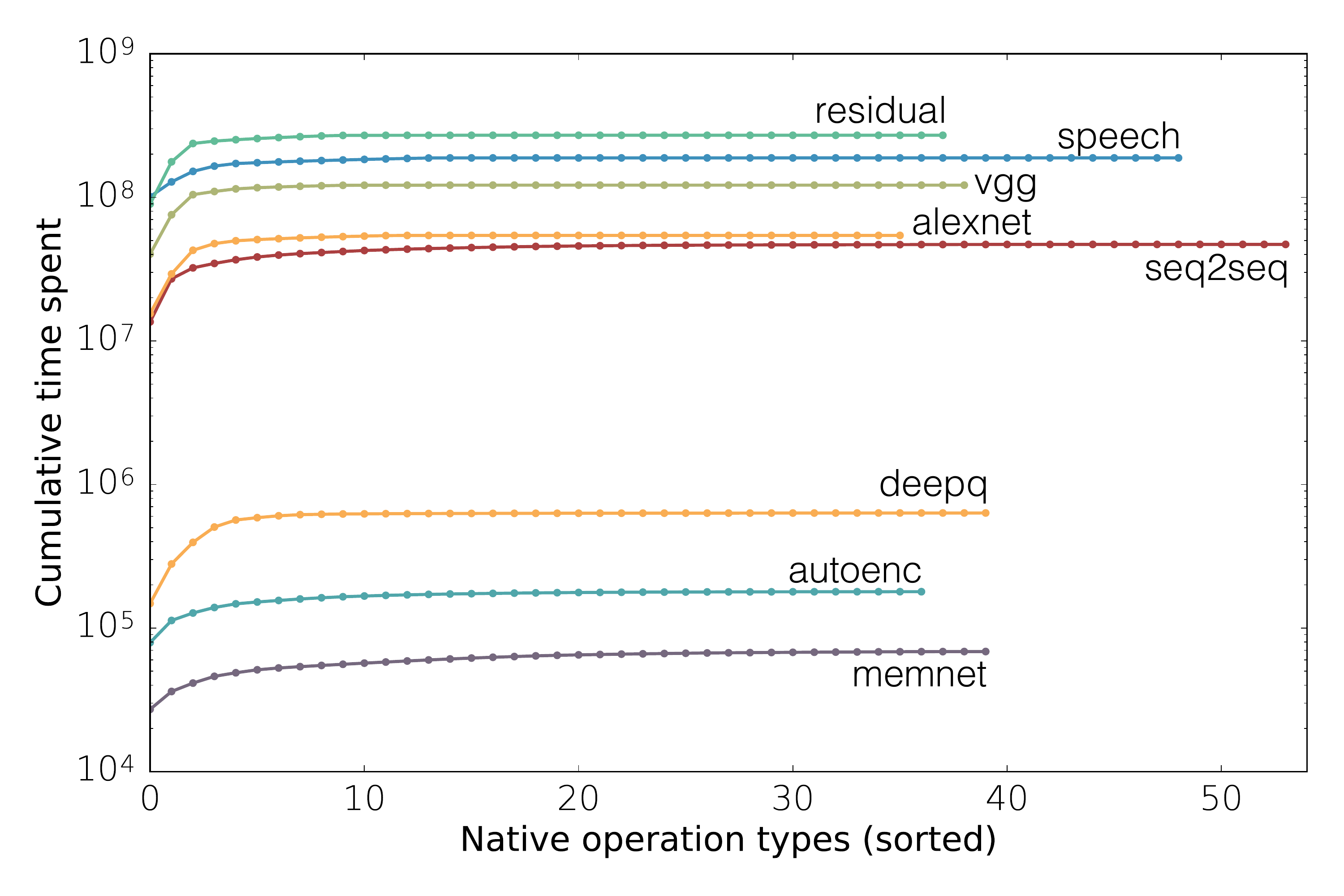}
\vspace{-2ex}
\caption{Total execution time is dominated by only a handful of unique operations.}
\label{fig:profiles}
\vspace{-3ex}
\end{center}
\end{figure}

\begin{figure*}[htbp]
\begin{center}
\includegraphics[width=0.8\textwidth]{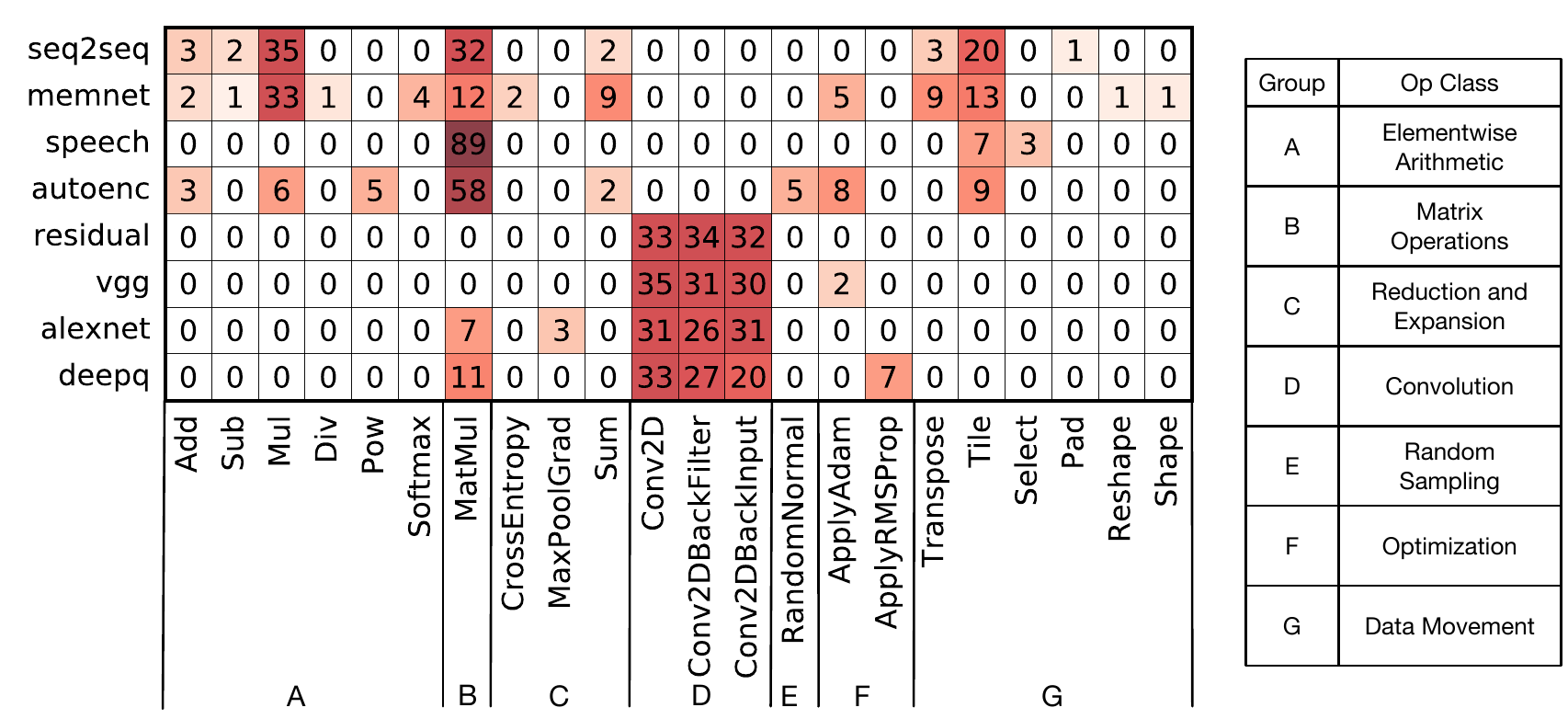}
\caption{Breakdown of execution time by operation type for each {\Name} workload.}
\label{fig:heatmap}
\vspace{-3ex}
\end{center}
\end{figure*}

\subsection{Operation type profiling}\label{subsec:ops}

The most basic performance question to ask about a workload is simply where the time is spent.
Many architects working on deep learning hardware already have some sense of this, but their viewpoints are often conflicting.
Some claim convolution, matrix-matrix multiplication, or matrix-vector multiplication are the predominant kernel for deep learning.
To an extent, they are all right, but the truth is somewhat more nuanced.
It depends on the model, environment, and use case being considered.

The general intuition about a handful of operation types dominating the overall computational time is true.
While it is an exaggeration to say that a workload can be reduced to a single operation, the distribution is quite skewed, as shown in Figure~\ref{fig:profiles}.
Each point on each curve represents the cumulative contribution to execution time from a single operation type.
It is clear that a handful of ``heavy'' operation types (usually 5 to 15) are collectively responsible for upwards of 90\% of the programs' duration.
It is important to note, however, that these types are not the same for every model (i.e., the leftmost point on each curve may represent a different operation), nor are they present in the same ratios or even at all.
Figure~\ref{fig:heatmap} shows a more detailed view of the time each model spends in a given operation type.
(For the sake of clarity, we only include operations with more than 1\% execution time, so a given row will sum to a value somewhere between 90\% and 100\%.)

Unsurprisingly, convolutional neural networks are indeed dominated by convolution, and fully-connected networks depend heavily on matrix multiplication.
On the other hand, the breakdown reveals a number of less-well understood trends.
For instance, while it is usually known that convolutional networks have gotten deeper and more expensive in recent years, it is usually not known that this has gone hand-in-hand with the gradual elimination of fully-connected layers.
Part of the justification for including \wkl{alexnet}, \wkl{vgg}, and \wkl{residual} is for exactly this kind of longitudinal comparison:
as the winners of the ILSVRC for 2012, 2014, and 2015 respectively, they share a dataset and machine learning task, and their structures are very similar.
However, \wkl{alexnet}'s two layers of locally-connected neurons constitute 11\% of its runtime, while \wkl{vgg}'s three fully-connected layers consume only 7\% and \wkl{residual}'s single fully-connected classification layer contributes less than 1\%.

We can also see the effects of intentional design trade-offs.
In Hannun et al.'s paper describing Deep Speech, the authors describe their decision to eschew more complicated components in favor of a structurally simple, easy-to-optimize networks:
\emph{``The complete RNN model...is considerably simpler than related models from the literature---we have limited ourselves to a single recurrent layer...and we do not use Long-Short-Term-Memory (LSTM) circuits.... By using a homogeneous model, we have made the computation of the recurrent activations as efficient as possible.''}
The evidence bears out their aims: \wkl{speech} is comprised almost exclusively of matrix-matrix multiplication operations, and the only other significant computations are part of the CTC loss function they require.

\subsection{Performance similarity}\label{subsec:similarity}

\begin{figure}[t]
\centering
\includegraphics[width=0.44\textwidth]{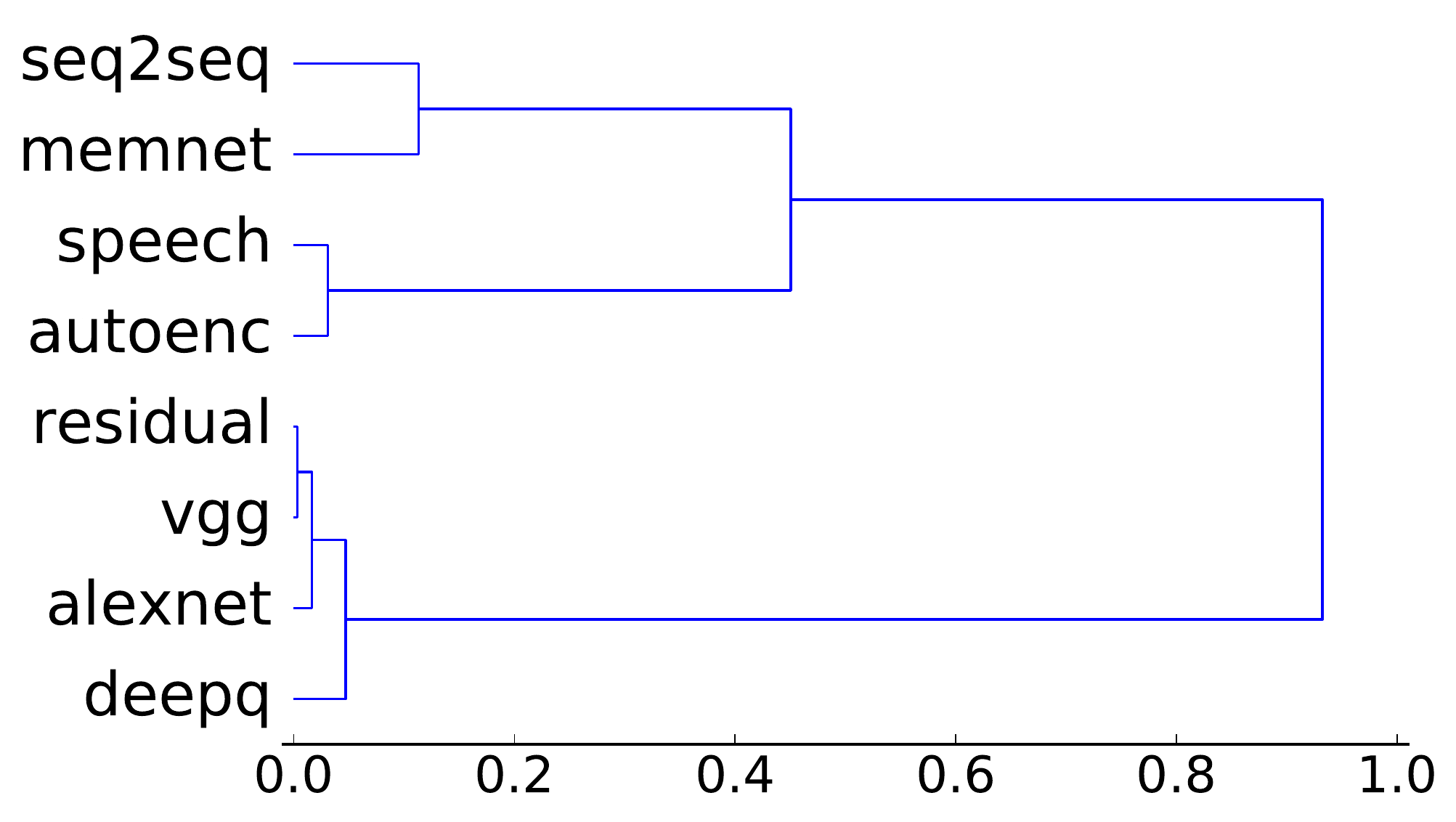}
\vspace{-1ex}
\caption{Hierarchical similarity in the {\Name} workloads. The tightly-clustered lower group includes all the convolutional networks.}
\vspace{-1ex}
\label{fig:dendrogram}
\end{figure}

Operation type profiling also offers a means of assessing similarity between workloads.
The mechanism is straightforward: each profile (a single row in Figure~\ref{fig:heatmap}) is interpreted as a vector in high-dimensional space.
Pairwise similarity can be computed using cosine similarity, and we use the inverse form
($1 - \frac{ \mathbf{A}\cdot\mathbf{B} }{ \lvert\mathbf{A}\rvert\lvert\mathbf{B}\rvert }$)
as a distance metric.
We can then use agglomerative clustering with centroidal linkage to understand their relationships---i.e., we greedily group the closest two vectors, compute their centroid, and repeat until we have a single, global cluster.
Figure~\ref{fig:dendrogram} presents a visual representation of the hierarchical clustering generated by this process.
The x-axis location of a linkage between two workloads (or two clusters of workloads) should be interpreted as a direct measure of the cosine distance between them, so the cosine distance between the centroid of the cluster containing \wkl{seq2seq} and \wkl{memnet} and that of  \wkl{speech} and \wkl{autoenc} is about 0.4.
High-dimensional distance is not an intuitive measure, so the value ``0.4'' is difficult to conceptualize, but relative distances can be understood in the normal sense: \wkl{speech} and \wkl{autoenc} have more similar performance profiles to each other than \wkl{seq2seq} and \wkl{memnet}.

To a deep learning expert, this dendrogram should be fairly unsurprising.
The three ImageNet challenge networks are grouped closely, and \wkl{deepq}, which relies heavily on convolutional layers, is not far off.
Somewhat less intuitive is the large distance between the two recurrent networks, \wkl{speech} and \wkl{seq2seq}.
This is not an artifact: it is a consequence of actual underlying dissimilarity.
While both networks are recurrent, Deep Speech uses CTC loss and a stack of fully-connected neurons, in contrast to the stateful LSTM neurons and standard cross entropy loss used by the sequence-to-sequence translation model.
The elementwise multiplications in \wkl{seq2seq} are a result of the LSTM neurons, and the data movement operations are part of the attention-based encoder/decoder it uses~\cite{bahdanau2015-attention_encoder}.

\subsection{Training and inference}\label{subsec:inference}

The vast majority of deep learning systems use some variant of gradient descent and backpropagation, which can be seen as two distinct phases:
in the forward phase, the model functions are used to compute an output value from a given input.
The model's parameters are fixed.
In the backward or update phase, the system evaluates its output on some criteria (a loss function).
Then, for every learnable parameter in the model, the system computes the partial derivative of the loss function with respect to that parameter.
This gradient is used to adjust the model parameters to minimize the value of the loss function.
Training a model requires evaluating both the forward and backward phases, while inference requires only the latter.

Architects working on deep learning systems generally understand that a rough symmetry exists between these two phases:
most functions evaluated in the forward phase have an analogue in the backwards phase with similar performance characteristics.
There are exceptions to this, however, such as the evaluation of the loss function, which is only computed during the backwards phase.
It is still a symmetric function (e.g., the softmax function and cross-entropy loss are duals), but both parts are evaluated only during training.
While this is not a revelation, it is easy to forget the performance implications of this fact.
Simple classifiers tend to have fast, cheap loss functions, but many deep learning models do not, and this can cause a skew in the both the overall performance of training with respect to inference as well as the relative importance of certain operation types.

\begin{figure}[htbp]
\begin{center}
\includegraphics[width=0.8\columnwidth]{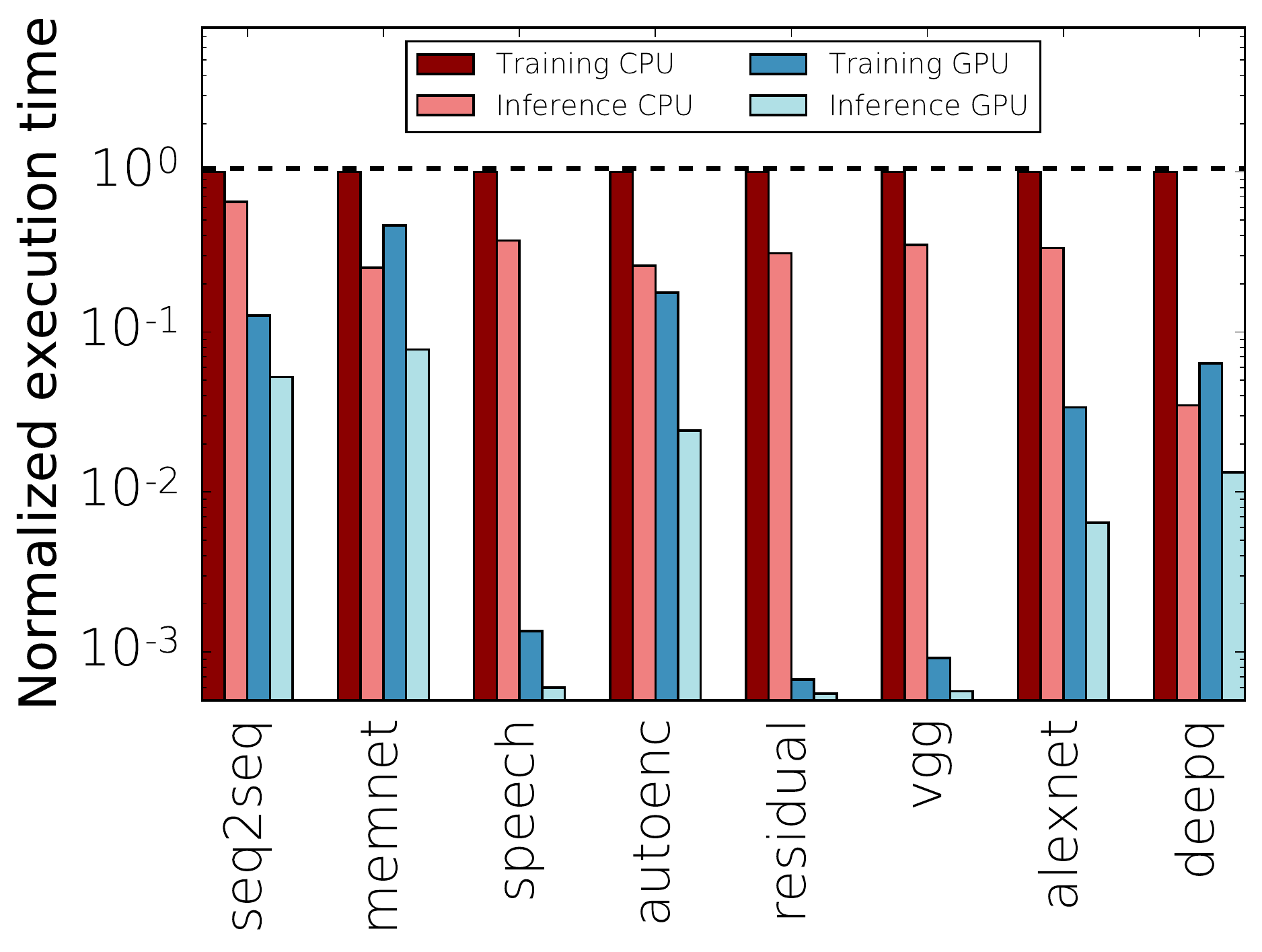}
\vspace{-2ex}%
\caption{The performance of training and inference relative to the training time of each Fathom on a CPU (i.e., the lowest performance configuration).
         Relative performance speedups indicate the benifits each workload has whether running on a CPU or GPU.}
\label{fig:inference}
\vspace{-2ex}%
\end{center}
\end{figure}

We show the normalized training and inference runtimes for all of the {\Name} workloads in Figure~\ref{fig:inference}.
Naturally, training time is more expensive than inference across the board, but the salient feature here is that it is variably faster.
Convolutional networks tend to pay a slightly higher cost for training because the convolutional partial gradient involves two reduction operations in the backwards phase (one for updating the filter weights and one for aggregating partials on the activity values) and only one in the forward phase.

\begin{figure*}[t]%
\centering%
\subfloat[Operation type scaling in \wkl{deepq}]{%
  \includegraphics[width=0.33\textwidth]{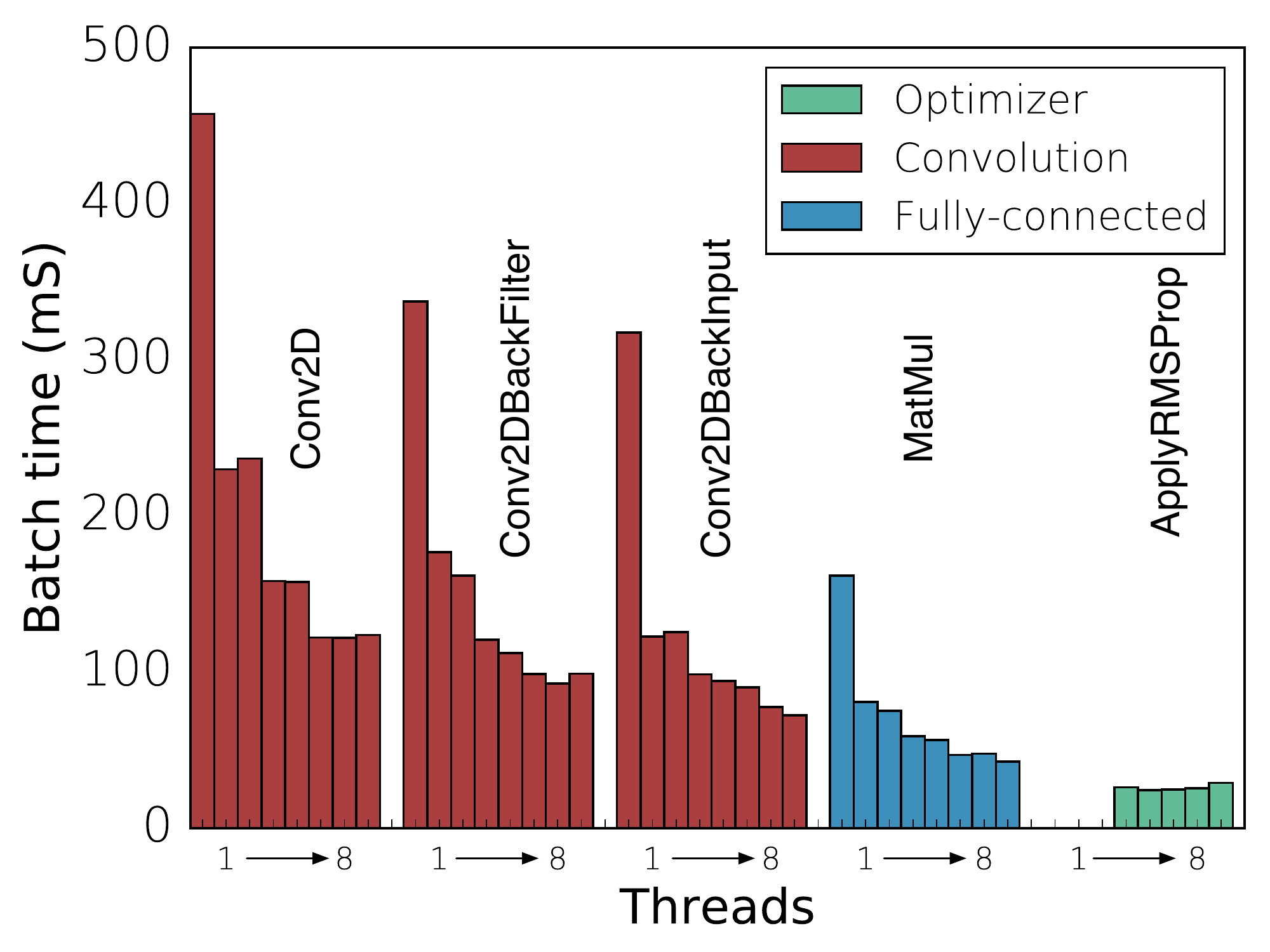}%
  \vspace{-1ex}%
  \label{fig:amdahl_atari}%
}\subfloat[Operation type scaling in \wkl{seq2seq}]{%
  \includegraphics[width=0.33\textwidth]{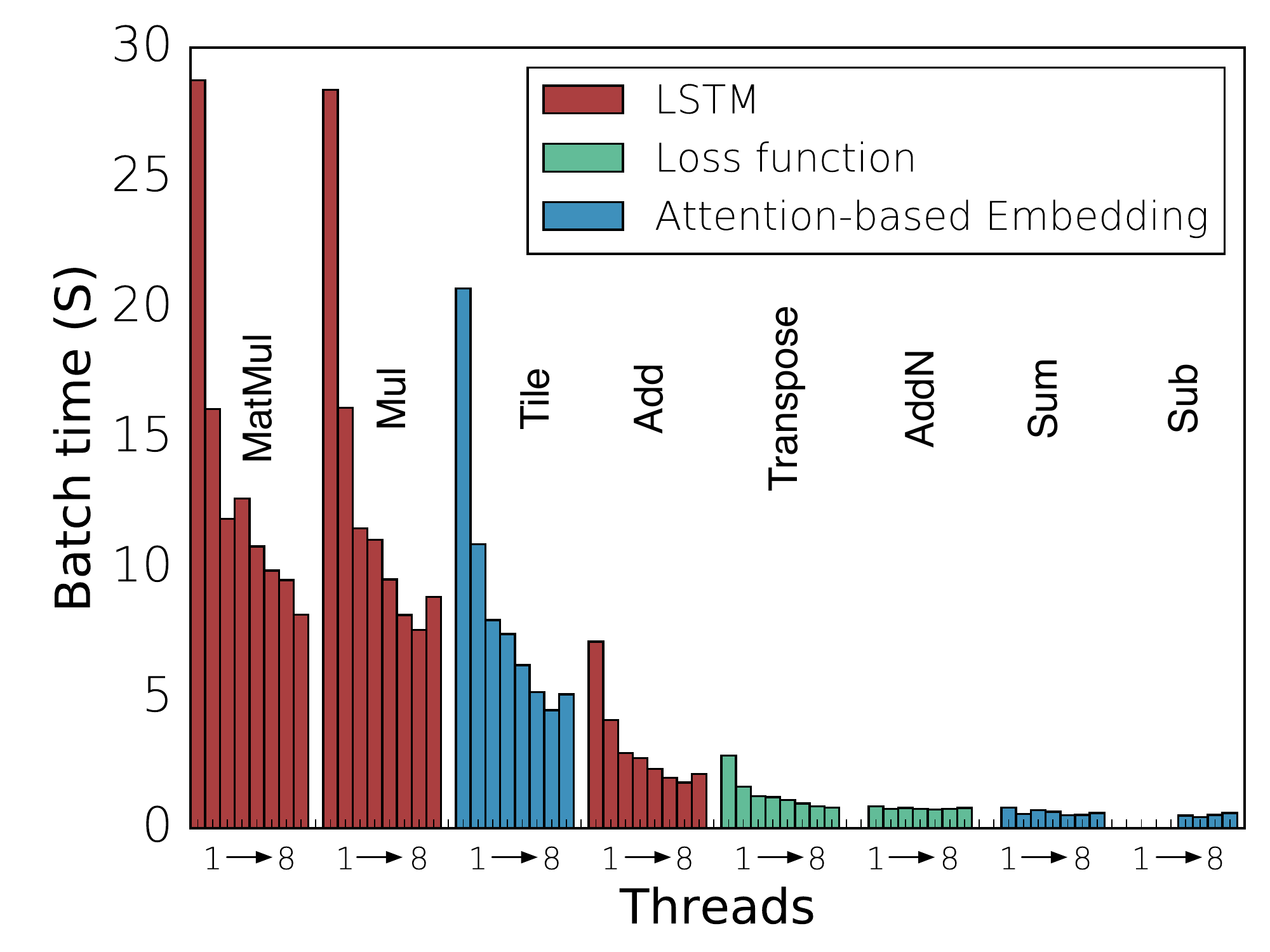}%
  \vspace{-1ex}%
  \label{fig:amdahl_seq2seq}%
}\subfloat[Operation type scaling in \wkl{memnet}]{%
  \includegraphics[width=0.33\textwidth]{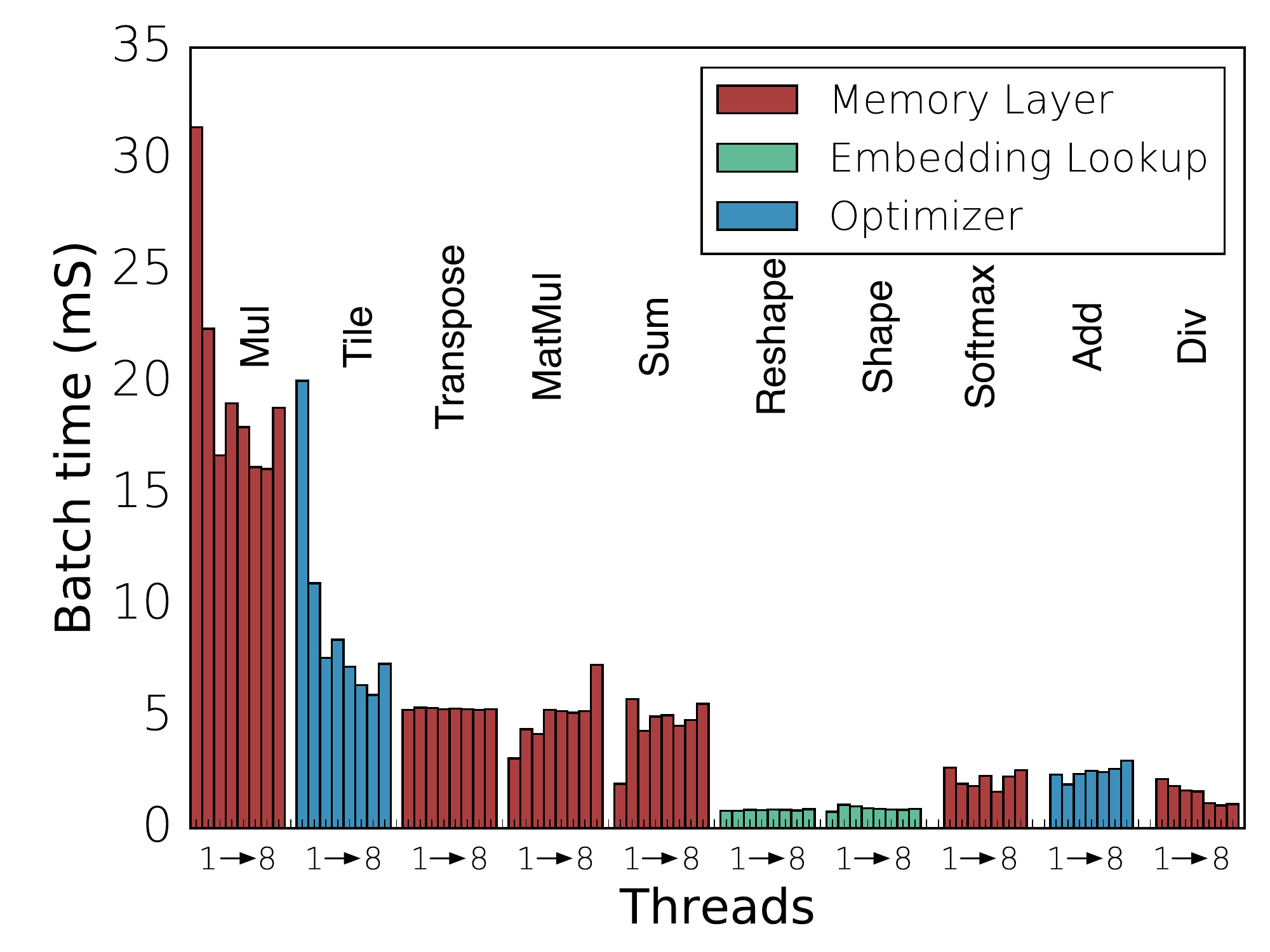}%
  \vspace{-1ex}%
  \label{fig:amdahl_memn2n}%
}%
\caption{The effect of Amdahl's law at the application level: the benefits of parallelizing matrix multiplication and convolution are limited by smaller, data-dependent operations in other parts of some models.}%
\vspace{-2ex}%
\label{fig:amdahl}%
\vspace{-1ex}%
\end{figure*}

We also evaluate both training and inference on a GPU for comparison\footnote{Running these experiments in aggregate avoids some of the detrimental effects of TensorFlow's CPU-only operations, as mentioned in Section~\ref{subsec:measurement}. They are still present, but the effect here is just an underestimate of the total speedup over a CPU, not erroneous conclusions.}.
As expected, GPU performance is substantially higher, especially on workloads with higher skew in their operation profile.
GPUs also experience variability in the train-to-inference ratio across workloads, and that variability tends to be strongly correlated with that of a CPU. 
That is, a large gap between training and inference times on the CPU implies a similar gap on the GPU.
The differences in absolute performance benefits can be partially attributed to the parallel scalability of the operations involved, which we examine in the next section.

\subsection{Parallelism and operation balance}\label{subsec:amdahl}

Many of the most dominant operations in deep learning workloads are amenable to parallelization, and nearly all of the existing work on architectural support for learning models involves parallel hardware to some extent.
The heaviest operations---convolution and matrix multiplication---are backed by an enormous body of prior work, and it is easy to overlook the effects of other operations.
The breakdown in Figure~\ref{fig:heatmap} revealed that while the distribution of operation types is skewed, the time spent in smaller operations is not zero.
As parallel resources are applied to the network, the operation types which scale strongly begin to diminish in relative importance, in accordance with Amdahl's law.
As a result, new operations types begin more important and the profile begins to flatten out.
We highlight three examples of this in Figure~\ref{fig:amdahl}.

Each of these plots shows the absolute time spent in each operation type as we increase the amount of parallelism available within an operation (using hooks in TensorFlow to specify the available thread pool for the underlying Eigen library).
We also annotate the operation types with their functional purpose in the deep learning model.
For instance, in Figure~\ref{fig:amdahl_atari}, the fourth cluster shows the time spent by \wkl{deepq} in matrix multiplication, which is the underlying operation behind the fourth and fifth fully-connected layers in that model.
In the extreme case of zero parallelism (single-threaded execution), the time spent in the convolutional and fully-connected parts of the model dominate the execution time to the point where no other operations even show up in the profile.
As parallelism increases to 8 threads, however, the optimizer (which has a large number of data-dependent operations) rises to around 7\% of the execution time (which is also seen in the heatmap of Figure~\ref{fig:heatmap}).
A similar effect is found in the \wkl{seq2seq} model: while the LSTM neurons and attention embedding portions of the model consume an enormous amount of time at low levels of parallelism, the loss function begins to become visible at higher thread counts.

The end-to-end memory model in Figure~\ref{fig:amdahl_memn2n} is a more complicated story.
Many of the operations in the memory layers operate on small, skinny tensors.
While these operations are frequent, they do not parallelize well (as the trip count is too low for thread-level parallelism, so the underlying library avoids it).
The elementwise multiplication is an exception (it operates on the final outputs of the memory layer, which is a wide tensor), as is the optimizer, which updates learning rates for a large collection of parameters simulatenously.

The lesson here is that the performance behavior of deep learning models is inextricably tied to their application-level structure.
While convolution and matrix multiplication are attractive targets for hardware support, there are limits to the benefits that can be extracted from them.
This is especially true for deep learning models with non-convolutional layers, sophisticated loss functions or optimization algorithms, or sparse storage.

%% file: related_work.tex

Other benchmark suites exist which do include deep learning algorithms.
The three most relevant are CortexSuite, BenchNN, and DjiNN and Tonic.
CortexSuite~\cite{cortex} is thematically based on the application areas surrounding perception and cognition, but it largely contains conventional algorithms.
BenchNN~\cite{chen2012-benchnn} uses machine learning methods to approximate the behavior of a subset of the PARSEC benchmarks.
While neural networks are used to some extent in BenchNN, its appeal was as a demonstration vehicle for the use of machine learning in approximate computing.
Modern deep learning models are significantly larger and more complex.
DjiNN and Tonic~\cite{gin} is perhaps the most similar benchmark suite.
The authors assemble several deep learning algorithms in order to study scalability and TCO of warehouse-scale computers, and several of their models are fairly modern.
Broadly, all three benchmarks seek to answer very different questions than {\Name}.
All three make no claims about representativeness, and many of the algorithms are now somewhat dated---most of the methods in {\Name} were published after these benchmarks were written.

Another relevant effort outside the academic literature are sets of sample codes commonly called ``model zoos''.
The most well-known of these is the Caffe Model Zoo~\cite{caffe-zoo}, though almost every popular deep learning framework has something similar.
A model zoo's primary purpose is as a sort of ``living documentation'' for the framework.
It showcases the correct use of library features in the context of popular deep learning methods, and it can serve as a gateway or tutorial for machine learning practitioners.
Model zoos and {\Name} both seek to implement a variety of state-of-the-art models, but they diverge in several key ways.
First, model zoos are often intentionally simplified implementations, written to highlight proper usage of the library as opposed to reflect realistic behavior.
Second, model zoos have no explicit goal to be representative of deep learning methods in general and sometimes have holes.
For instance, at the time of writing, Caffe had no support for CTC optimization and LSTM support was fairly new, so its model zoo included only experimental recurrent neural networks.
Finally, and perhaps most importantly for researchers, model zoos have no standard interface.
The networks in a model zoo are normally user-contributed, and they rarely look remotely alike.
This poses a logistical challenge for a researchers attempting to evaluate the effects of a hardware change on a battery of models in a consistent manner.
All {\Name} models are wrapped in a standard interface which exposes the same functions for every model.
Thus, evaluating training, inference, or simply inspecting the model's dataflow graph is straightforward.

With respect to high-level performance analysis tools for deep learning, there are few examples, primarily because deep learning frameworks are still a fairly recent phenomenon.
The two most prominent exceptions are TensorBoard and EEG, both Google-developed tools for TensorFlow~\cite{tensorflow}.
TensorBoard is a visualization tool for TensorFlow's dataflow graphs, but it can also display parameter values and limited timing information.
EEG is a distributed tracing tool which can reconstruct the dynamic execution timeline of TensorFlow operations, even over separate devices.
Unfortunately, Google has not released EEG to the public.

%% file: conclusion.tex

This paper presents the Fathom deep learning workloads.
Fathom is intended to establish a standard for researchers to quantitatively develop better hardware and systems for deep learning. Fathom alleviates the burden of workload selection by assembling eight modern deep learning models into a single, unified package with a consistent interface.
We also provide insights including how similar the eight different models are, where the cycles are being spent, and how optimizations (e.g., parallel execution) shift performance bottlenecks, guiding researcher's direction to focus on the important problems.
As the field continues to evolve, there will inevitably be new models which arise, and we hope Fathom will become a ``living'' workload suite, incorporating advances as they are discovered.